# Outlier Detection Using Vector Cosine Similarity by Adding a Dimension

Zhongyang Shen

*Abstract*— We propose a new outlier detection method for multi-dimensional data. The method detects outliers based on vector cosine similarity, using a new dataset constructed by adding a dimension with zero values to the original data. When a point in the new dataset is selected as the measured point, an observation point is created as the origin, differing only in the new dimension by having a non-zero value compared to the measured point. Vectors are then formed from the observation point to the measured point and to other points in the dataset. By comparing the cosine similarities of these vectors, abnormal data can be identified. An optimized implementation (MDOD) is available on PyPI: https://pypi.org/project/mdod/.

*Keywords*— outlier detection, multi-dimensional data, vector cosine similarity

## I. Introduction

Outliers show their special characteristics in certain situations, the identification of outliers, is an important topic in data processing. At present, there are many methods such as LOF, Isolation Forest, OGAD [1], ABOD [3], DBSCAN [4] and other algorithms to identify outliers in two-dimensional or three-dimensional datasets. In some data application cases, we will encounter requirements for identifying outliers in high-dimensional data. For high-dimensional data, as the number of dimensions increases, the number of calculations will increase exponentially, which poses challenges to finding the correlation of high-dimensional features in outlier detection. Some algorithms, such as dimension reduction like PCA algorithm, have led to the loss of multidimensional data information and some incorrect results. At present, there are ABOD and other algorithms that support high-dimensional data outlier detection, but they have their own shortcomings in terms of computational complexity, training data, and scale settings.

We propose an unsupervised high-dimensional data outlier detection algorithm, Outliers Detected by Adding a Dimension to Compare Vector Cosine Similarity (OD-ADVCS). Using the algorithm, we can solve the problem of computational complexity caused by the increase in dimensionality. The algorithm detects outliers based on vector cosine similarity: first, we add one new dimension to the original data and assign zero-value to the new dimension; then, each point in the dataset is selected as a measured point, and a new observation point is created that is the same as the measured point except for the new dimension assigned to non-zero value. The observation point is assigned as the origin, the vector from observation point to measured point and the vector from observation point to the other point in the new (n+1)-dimensional dataset are formed, then we calculate and compare the vector cosine similarity between the formed vectors to filter out abnormal data. Experiments show that the method can effectively detect outliers in datasets of different high-dimensional types.

## II. Algorithm

### A. Original dataset

We define the quantity of n-dimensional dataset q, and the points are marked as $X_i$ ($m_1, m_2,...,m_n$), where $X_i$ is the data point and i⩽q, n is the dimension of the data point, and $m_n$ is the n-th dimension value of the $X_i$ point.

Thus, $X_i(m_k)$ is marked as the value of the k-th dimension of the point i, i⩽q, k⩽n, and $X_j(m_k)$ is marked as the value of the k-th dimension of the point j, j⩽q, k⩽n.

### B. Principle

Outliers are detected by the following steps.

- Step 1: For a given n-dimensional dataset with a given quantity of dataset points q(q > 2, n ⩾ 2), each n-dimensional point $X_i(m_1,m_2,...,m_n)$ is extended by one dimensional to (n+1)-dimensional $X_i(m_1,m_2,...,m_n,m_{n+1})$, and the value of the new dimensional $m_{n+1}$ is assigned to 0.

- Step 2: As a measured point $X_i(m_1,m_2,...,m_n,m_{n+1})$ selected from the (n+1)-dimensional data points in the dataset p respectively, we create a new observation point $O_i(m_1,m_2,...,m_n,m_{n+1})$ with (n+1)-dimension, whose value is the same as the measured point $X_i(m_1,m_2,...,m_n,m_{n+1})$ in each dimension, except the value of the (n+1)th dimension is set to a non-zero value, which is different from the measured point. Thus, for the origin of the new observation point $O_i$, the values of every dimension are the same as $X_i$, except the (n+1)-dimension, with a difference: $O_i$ ($m_{n+1}$) ≠ 0, $X_i$ ($m_{n+1}$)=0.

- Step 3: Take the observation point $O_i(m_1,m_2,...,m_n,m_{n+1})$ as the origin, and a vector $O_i \rightarrow X_i$ to the measured point $X_i(m_1,m_2,...,m_n,m_{n+1})$ is formed.

- Step 4: Another new vector $O_i \rightarrow X_j$ is formed from the observation point $O_i(m_1,m_2,...,m_n,m_{n+1})$ to the other reference point $X_j(m_1,m_2,...,m_n,m_{n+1})$ in the dataset.
- Step 5: Calculate the cosine similarity between the vector $O_i \rightarrow X_i$ and the vector $O_i \rightarrow X_j$. The cosine similarity $S_{ij}$ between the vector $O_i \rightarrow X_i$ and the vector $O_i \rightarrow X_j$ ($i \neq j, i \leq q, j \leq q$) is calculated.

$$S_{ij} = \frac{\sum_{k=1}^{n+1}(|O_i(m_k) - X_i(m_k)| * |O_i(m_k) - X_j(m_k)|)}{\sqrt{\sum_{k=1}^{n+1}(O_i(m_k) - X_i(m_k))^2} * \sqrt{\sum_{k=1}^{n+1}(O_i(m_k) - X_j(m_k))^2}}$$

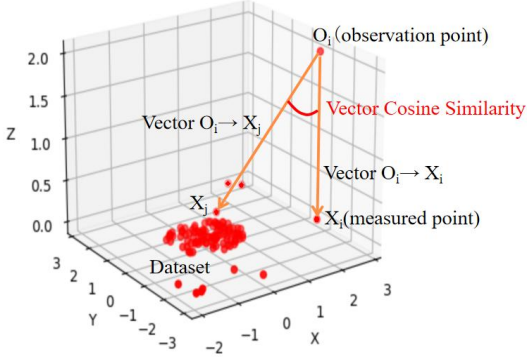

Figure 1. Example of Outlier Detection Using Vector Cosine Similarity by Adding a Dimension

- Step 6: Repeat Step 4-5 to calculate the vector cosine similarity corresponding to all other reference points in dataset except the measured point.
- Step 7: Summing the largest r value of cosine similarity of vectors as the anomaly calculated value of this measured point. Sorting the maximum first r values of $S_{ij}$ and summation as $SUM_i$ ($r \leq q$). Thus, we obtain the cosine similarity value of point i.
- Step 8: Repeat Step 2-7, take the next point as the measured point $X_i(m_1,m_2,...,m_n,m_{n+1})$, until all m data are measured. Repeat the calculation of the next point until all points have been calculated.
- Step 9: Comparing the vector cosine similarity anomaly calculation values of all m points, sorting $SUM_i$, the smaller the value, the more likely it is an outlier. The smaller the value is, the more it tends to be an outlier value; otherwise, the larger the value, the more it is like a normal point.

## III. PSEUDO-CODE

The following pseudo code is based on the thought of algorithm described above.

---
Algorithm Program
1: input original dataset $X_i(m_1,m_2,...m_n)$, $i \in q$
2: new expanded dateset $X_i(m_1,m_2,...m_n,m_{n+1})$, $m_{n+1}=0$, $i \in q$
3: **for** $i=1 \rightarrow q$
4:   observation point created $O_i(m_1,m_2,...m_n,m_{n+1})$, $m_{n+1} \neq 0$
5:   **for** $j=1 \rightarrow q, j \neq i$
6:   $$S_{ij} = \frac{\sum_{k=1}^{n+1}(|O_i(m_k) - X_i(m_k)| * |O_i(m_k) - X_j(m_k)|)}{\sqrt{\sum_{k=1}^{n+1}(O_i(m_k) - X_i(m_k))^2} * \sqrt{\sum_{k=1}^{n+1}(O_i(m_k) - X_j(m_k))^2}}$$
7:   $SortS_i(r) = Sorting(S_{ij}, Max \rightarrow Min)$
8:   $SUM_i = \sum_{k=1}^{r(r \leq q)} SortS_i(r)$
9:   **end for**
10:  $SortSum(t) = Sorting(SUM_i, Min \rightarrow Max)$
11: **end for**
12: Outlier $\leftarrow$ Point(SortSum(t)), Min $\rightarrow$ Max
---

## IV. EXPERIMENTAL DATA AND ANALYSIS

The current conventional outlier recognition algorithms, LOF for 2D data and ABOD for 3D data and high-dimensional data, will be used as comparisons with the new method in this paper.

There are two major parameters used in OD-ADVCS: $n_d$, the value of the new dimension of the observation point, and $s_n$, the largest static number of cosine similarity of the vector's value.

- 2D and 3D test data are formed by using python random method to generate random numbers with normal value cluster range and anomalies according to the specified shape designed by the requirements to compare the algorithms. Below, Table I shows the 3D test data statistical result of the new algorithm compared with algorithm ABOD, and Table II shows the 2D test data statistical result of the new algorithm compared with algorithm LOF.
- Clover flower data are used as test data to detect outliers by OD-ADVCS, compared with algorithm ABOD. The clover flower dataset comes from the UCI 4-dimensional dataset with three sets. Datasets are formed by normal data selected from one set and abnormal data selected from other two different sets. According to the marked classification, we compare the accuracy of identification between the new algorithm and the common algorithm ABOD. Following Table III(a), which shows the result when one outlier is selected, Table III(b) shows the result when two outliers are selected. In this experiment, the data is first normalized, and then multiplied by 300 to enlarge the differentiation to facilitate data comparison.

Table I. Experiments: 3D test data

| Normal Data QTY | Normal Points Radius Range | Abnormal Data QTY | Abnormal Points Distribution Radius Range | Test Times | OD-ADVCS | | | | ABOD | |
|---|---|---|---|---|---|---|---|---|---|---|
| | | | | | $n_d$ | $s_n$ | Accurate Recognition Times | Accuracy Recognition Rate | Accurate Recognition Times | Accurate Recognition Times |
| 200 | 1R | 20 | 1.10R-3R | 200 | 80 | 40 | 192 | 96% | 166 | 83.0% |
| 200 | 1R | 20 | 1.20R-3R | 200 | 80 | 40 | 200 | 100% | 181 | 90.5% |
| 200 | 1R | 20 | 1.20R-3R | 200 | 80 | 10 | 198 | 99.0% | | |
| 200 | 1R | 20 | 1.30R-3R | 200 | 80 | 10 | 200 | 100% | 194 | 97.0% |
| 215 | 1R | 5 | 1.10R-3R | 200 | 80 | 40 | 200 | 100% | 196 | 98.0% |
| 215 | 1R | 5 | 1.20R-3R | 200 | 80 | 40 | 200 | 100% | 200 | 100% |
| 215 | 1R | 5 | 1.20R-3R | 200 | 80 | 10 | 200 | 100% | | |
| 215 | 1R | 5 | 1.30R-3R | 200 | 80 | 10 | 200 | 100% | 200 | 100% |

Table II. Experiments: 2D test data

| Normal Data QTY | Normal Points Radius Range | Abnormal Data QTY | Abnormal Points Distribution Radius Range | Test Times | OD-ADVCS | | | | LOF | |
|---|---|---|---|---|---|---|---|---|---|---|
| | | | | | $n_d$ | $s_n$ | Accurate Recognition Times | Accuracy Recognition Rate | Accurate Recognition Times | Accurate Recognition Times |
| 200 | 1R | 20 | 1.10R-3R | 200 | 80 | 40 | 194 | 97% | 186 | 93% |
| 200 | 1R | 20 | 1.20R-3R | 200 | 80 | 40 | 200 | 100% | 193 | 96.5% |
| 200 | 1R | 20 | 1.20R-3R | 200 | 80 | 10 | 199 | 99.5% | | |
| 200 | 1R | 20 | 1.30R-3R | 200 | 80 | 10 | 200 | 100% | 200 | 100% |
| 215 | 1R | 5 | 1.10R-3R | 200 | 80 | 40 | 200 | 100% | 196 | 98% |
| 215 | 1R | 5 | 1.20R-3R | 200 | 80 | 40 | 200 | 100% | 199 | 99.5% |
| 215 | 1R | 5 | 1.20R-3R | 200 | 80 | 10 | 200 | 100% | | |
| 215 | 1R | 5 | 1.30R-3R | 200 | 80 | 10 | 200 | 100% | 200 | 100% |

Table III(a). Experiments: Clover flower with 4-dimensional data when one outlier is selected

| Normal Data | Abnormal Data | Normal Data QTY | Abnormal Data QTY | Test Times | OD-ADVCS | | ABOD | |
|---|---|---|---|---|---|---|---|---|
| | | | | | Accurate Number of Times | Accuracy Rate | Accurate Number of Times | Accuracy Rate |
| Iris-setosa | Iris-versicolor | 48 | 1 | 50 | 50 | 100.0% | 49 | 98.0% |
| | Iris-virginica | 48 | 1 | 49 | 49 | 100.0% | 49 | 100.0% |
| Iris-versicolor | Iris-setosa | 50 | 1 | 48 | 48 | 100.0% | 42 | 87.5% |
| | Iris-virginica | 50 | 1 | 49 | 35 | 71.4% | 23 | 46.9% |
| Iris-virginica | Iris-setosa | 49 | 1 | 48 | 48 | 100.0% | 31 | 64.6% |
| | Iris-versicolor | 49 | 1 | 50 | 15 | 30.0% | 14 | 28.0% |
| Summary | | | | 294 | 245 | 83.3% | 208 | 70.7% |

Table III(b). Experiments:Clover flower with 4-dimensional data when two outliers are selected

| Normal Data | Abnormal Data | Normal Data QTY | Abnormal Data QTY | Test Times | OD-ADVCS | | ABOD | |
|---|---|---|---|---|---|---|---|---|
| | | | | | Accurate Number of Times | Accuracy Rate | Accurate Number of Times | Accuracy Rate |
| Iris-setosa | 2 Iris-versicolor | 48 | 2 | 1225 | 1225 | 100.0% | 528 | 43.1% |
| | 2 Iris-virginica | 48 | 2 | 1176 | 1176 | 100.0% | 1035 | 88.0% |
| | 1 Iris-versicolor + 1 Iris-virginica | 48 | 2 | 2450 | 2450 | 100.0% | 2209 | 90.2% |
| Iris-versicolor | 2 Iris-setosa | 50 | 2 | 1128 | 1128 | 100.0% | 190 | 16.8% |
| | 2 Iris-virginica | 50 | 2 | 1176 | 553 | 47.0% | 190 | 16.2% |
| | 1 Iris-setosa + 1 Iris-virginica | 50 | 2 | 2352 | 1680 | 71.4% | 966 | 41.1% |
| Iris-virginica | 2 Iris-setosa | 49 | 2 | 1128 | 1128 | 100.0% | 1 | 0.1% |
| | 2 Iris-versicolor | 49 | 2 | 1225 | 70 | 5.7% | 66 | 5.4% |
| | 1 Iris-setosa + 1 Iris-versicolor | 49 | 2 | 2400 | 720 | 30.0% | 390 | 16.3% |
| Summary | | | | 14260 | 10130 | 71.0% | 5575 | 39.1% |

- Wilt dataset with 6-dimensional dimension are used as test data to detect outliers by OD-ADVCS, compared with algorithm ABOD. Wilt dataset comes from LMU Dataset Wilt (2% of outliers version#08, Normalized, duplicates). The dataset is formed by 93 wilt data and 4578 non-wilt data, with 2% wilt data among the total of 4671 data. Following Table IV shows the result of the accuracy of identification by the algorithm OD-ADVCS. Because the smaller the calculated score value, the more outliers tend to be, and the evaluation method we use here is to count the number of outliers in each percentage area sorted by the minimum value rank. By comparing the statistical number of outliers in each percentage range, we evaluate the effectiveness of the algorithm. In this experiment the parameter $n_d$ is set to 200, and $s_n$ is set to 15.

Table IV. 6-dimensional data result accuracy by OD-ADVCS

| Score Rank Scope | Data Ranking | Qty | Wilt Qty | Wilt in Total | Wilt in Scope | Identified Rate Increment |
|---|---|---|---|---|---|---|
| 0-1% | 1-47 | 47 | 33 | 35.5% | 70.2% | 35.5% |
| 1-2% | 48-93 | 46 | 15 | 16.1% | 32.6% | 51.6% |
| 2-3% | 94-140 | 47 | 13 | 14.0% | 27.7% | 65.6% |
| 3-4% | 141-186 | 46 | 12 | 12.9% | 26.1% | 78.5% |
| 4-5% | 187-233 | 47 | 7 | 7.5% | 14.9% | 86.0% |
| 5-6% | 234-279 | 46 | 6 | 6.5% | 13.0% | 92.5% |
| 6-7% | 280-326 | 47 | 7 | 7.5% | 15.1% | 100% |
| SUM | - | | 93 | 100% | | - |

- We try to compare the influence of different parameter values of $n_d$ and $s_n$, with the same Wilt dataset used in the experiment, Figure 2 and Figure 3 show the trends. The evaluation method we use here is to record the most deviated score ranking, that is, the largest ranking, that has been marked as an outlier. By comparing this ranking, we evaluate the influence of the parameters of the algorithm.

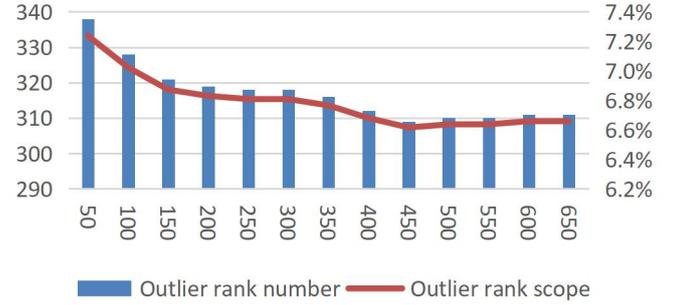

Figure 2. Influence of value of $n_d$ (new dimension of observation point)

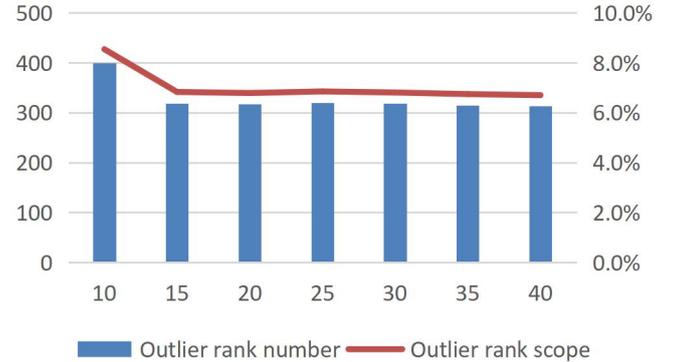

Figure 3. Influence of $s_n$ (largest static number of vector cosine similarity)

According to the data analysis, we find that the OD-ADVCS algorithm has strong accuracy and adaptability for identifying data outliers in each dimension.

- From the experimental data, it shows that the OD-ADVCS algorithm has the same recognition accuracy as the traditional two- or three-dimensional algorithm LOF and the angle-based algorithms.
- OD-ADVCS has obvious advantages in high-dimensional datasets.
- It shows from the experiment that the parameter selection of the OD-ADVCS algorithm refers to a setting with weak sensitivity to the results and strong adaptability.

An optimized Python implementation of the proposed OD-ADVCS algorithm, named MDOD (Multi-Dimensional Outlier Detection), is publicly available. The library leverages vectorized NumPy operations and scikit-learn's NearestNeighbors to efficiently compute the top-r similarities, significantly improving scalability on larger datasets while maintaining mathematical equivalence to the original formulation. It can be easily installed via PyPI using the command 'pip install mdod' (https://pypi.org/project/mdod/).

V. Conclusion

Based on vector cosine similarity, we provide a method OD-ADVCS for finding outliers by adding one dimensional to the dataset. Experiments reveal that the recognition accuracy of algorithm OD-ADVCS is similar to that of the two-dimensional or three-dimensional algorithms, it also has larger advantages in multi-dimensional data accuracy and adaptability.

References

[1] Zhongyang Shen, "Outlier Geometric Angle Detection Algorithm," 2019 International Conference on Artificial Intelligence in Information and Communication (ICAIIC), 2019, pp. 316-321, doi: 10.1109/ICAIIC.2019.8669090.

[2] Zhongyang Shen, "Three-dimensional Data Outlier Detected by Angle Analysis," 2022 International Conference on Artificial Intelligence in Information and Communication (ICAIIC), Jeju Island, Korea, Republic of, 2022, pp. 446-450, doi: 10.1109/ICAIIC54071.2022.9722637.

[3] Hans-Peter Kriegel, Matthias Schubert, Arthur Zimek, "Angle-Based Outlier Detection in High-dimensional Data," The 14th ACM SIGKDD International Conference on Knowledge Discovery and Data Mining (KDD), 2008.

[4] Martin Ester, Hans-Peter Kriegel, Jörg Sander, Xiaowei Xu, "A density-based algorithm for discovering clusters in large spatial databases with noise," Proceedings of the Second International Conference on Knowledge Discovery and Data Mining (KDD),AAAI Press.,1996.

[5] Fisher,R. A.. (1988). Iris. UCI Machine Learning Repository. https://doi.org/10.24432/C56C76.

[6] https://www.dbs.ifi.lmu.de/research/outlier-evaluation/DAMI/semantic/Wilt/Wilt_02_v08.html#Wilt_norm_02_v08.

[7] https://scikit-learn.org/stable/modules/metrics.html#cosine-similarity.

[8] Zhongyang Shen, "Cluster Quantity Distinguished by Geometric Angle Measurement," 2020 International Conference on Artificial Intelligence in Information and Communication (ICAIIC), 2020, pp. 514-519, doi: 10.1109/ICAIIC48513.2020.9065253.

[9] David Arthur, Sergei Vassilvitskii, "k-means++: the advantages of careful seeding," Proceedings of the eighteenth annual ACM-SIAM symposium on Discrete algorithms, 2007, pp 1027–1035.

[10] Xie Huajuan, "Unsupervised Learning Methods and Applications," Publishing House of Electronics Industry, China, 2016.

[11] Masashi Sugiyama, "An Illustrated Guide to Machine Learning," Kodansha Ltd., Japan, 2013.

[12] Toby Segaran, "Programming Collective Intelligence," O' Reilly Media, Inc., 2007.

[13] Pankaj K. Agarwal and Nabil H. Mustafa, "k-means projective clustering," In PODS'04:Proceedings of the twenty-third ACM SIGMODSIGACT-SIGART symposium on Principles of database systems, pages 155–165, ACM, 2004.

[14] Koki Saitoh, "Deep Learning from Scratch," O' Reilly Japan, Inc., 2016

[15] Tapas Kanungo, David M. Mount, Nathan S. Netanyahu, Christine D. Piatko, Ruth Silverman, and Angela Y. Wu, "A local search approximation algorithm for k-means clustering," Proceedings of the eighteenth annual symposium on Computational geometry, ACM, 2002, DOI:https://doi.org/10.1145/513400.513402.

[16] Bardia Yousefi and Chu Kiong Loo,"Comparative study on interaction of form and motion processing streams by applying two different classifiers in mechanism for recognition of biological movement," The Scientific World Journal, 2014.

[17] Andreas C. Muller, Sarah Guido, "Introduction to machine Learning with Python," O' Reilly Media, Inc., 2016.

[18] Zhou Zhihua, "Machine Learning," Tsinghua University Press, 2016.